\documentclass[final,3p,times,twocolumn]{elsarticle}

\makeatletter
\def\ps@pprintTitle{%
 \let\@oddhead\@empty
 \let\@evenhead\@empty
 \def\@oddfoot{}%
 \let\@evenfoot\@oddfoot}
\makeatother

\usepackage{amssymb}
\usepackage{hyperref}
\usepackage{xspace}
\usepackage{graphicx}
\usepackage{xcolor}
\usepackage{rotating}
\usepackage{subcaption}
\newcommand{\myparagraph}[1]{\smallskip \noindent \textbf{#1}}

\newcommand{\art}{\textit{ART}\xspace}
\newcommand{\cleverhans}{\textit{CleverHans}\xspace}
\newcommand{\foolbox}{\textit{Foolbox}\xspace}
\newcommand{\pytorch}{\textit{PyTorch}\xspace}
\newcommand{\sklearn}{\textit{scikit-learn}\xspace}
\newcommand{\keras}{\textit{Keras}\xspace}
\newcommand{\tensorflow}{\textit{TensorFlow}\xspace}
\newcommand{\matplotlib}{\textit{matplotlib}\xspace}
\newcommand{\torchvision}{\textit{torchvision}\xspace}
\newcommand{\numpy}{\textit{numpy}\xspace}
\newcommand{\scipy}{\textit{scipy}\xspace}

\newcommand{\adv}{\texttt{adv}\xspace}
\newcommand{\ml}{\texttt{ml}\xspace}
\newcommand{\explanation}{\texttt{explanation}\xspace}
\newcommand{\optim}{\texttt{optim}\xspace}
\newcommand{\data}{\texttt{data}\xspace}
\newcommand{\pkfigure}{\texttt{figure}\xspace}
\newcommand{\secml}{\texttt{secml}\xspace}

\newcommand{\xmark}{\texttt{x}\xspace}

\begin{document}

\begin{frontmatter}



\title{\secml: Secure and Explainable Machine Learning in Python}


\author[1,2]{Maura Pintor}
\author[1,2]{Luca Demetrio}
\author[1,2]{Angelo Sotgiu}
\author[1]{Marco Melis}
\author[1]{Ambra Demontis}
\author[1,2]{Battista Biggio}

\address[1]{DIEE, University of Cagliari, Via Marengo, Cagliari (IT)}
\address[2]{Pluribus One, Via Vincenzo Bellini, 9, Cagliari (IT)}

\begin{abstract}
We present \secml, an open-source Python library for secure and explainable machine learning. 
It implements the most popular attacks against machine learning, including test-time evasion attacks to generate adversarial examples against deep neural networks and training-time poisoning attacks against support vector machines and many other algorithms.
These attacks enable evaluating the security of learning algorithms and the corresponding defenses under both white-box and black-box threat models.
To this end, \secml provides built-in functions to compute security evaluation curves, showing how quickly classification performance decreases against increasing adversarial perturbations of the input data.
\secml also includes explainability methods to help understand why adversarial attacks succeed against a given model, by visualizing the most influential features and training prototypes contributing to each decision.
It is distributed under the Apache License 2.0 and hosted at \url{https://github.com/pralab/secml}. 
\end{abstract}


\end{frontmatter}

%\linenumbers
\section{Introduction}
\label{sec:intro}

Machine learning is vulnerable to well-crafted attacks.
At test time, the attacker can stage evasion attacks (i.e., adversarial examples)~\cite{huang11,biggio13-ecml,szegedy14-iclr,papernot16-sp,carlini17-sp}, sponge examples~\cite{shumailov2021sponge}, model stealing~\cite{tramer2016stealing}, and membership inference~\cite{shokri2017membership} attacks to violate system integrity, availability, or even its confidentiality.
Similarly, at training time, the attacker may target either system integrity or availability via poisoning~\cite{biggio12-icml, biggio18} and backdoor attacks\cite{gu2019badnets}. 
The most studied attacks, namely evasion and poisoning, firstly explored by Biggio et al.~\citep{biggio12-icml,biggio13-ecml}, are formalized as constrained optimization problems, solved through gradient-based or gradient-free algorithms~\citep{biggio18,joseph18-advml-book}, depending on whether the attacker has white- or black-box access to the target system. Many libraries implement the former, however, they do not allow developers to assess machine learning models' security easily. Hence, we present \secml, an open-source Python library that serves as a complete tool for evaluating and assessing the performance and robustness of machine-learning models.
To this end, \secml implements:
($i$) a methodology for the empirical security evaluation of machine-learning algorithms under different evasion and poisoning attack scenarios; and
($ii$) explainable methods to help understand why and how these attacks are successful. 
With respect to other popular libraries that implement attacks almost solely against Deep Neural Networks (DNNs)~\citep{papernot18-technical_report,rauber17-arxiv,art18}, \secml also implements training-time poisoning attacks and computationally-efficient test-time evasion attacks against many different algorithms, including support vector machines (SVMs) and random forests (RFs).
It also incorporates both the feature-based and prototype-based explanation methods proposed by~\cite{ribeiro16,sundararajan17-icml,koh2017understanding}. 

\begin{figure*}[t]  
    \begin{subfigure}{0.165\textwidth}
        \vspace{7pt}
        \resizebox{\textwidth}{!}{%
        \begin{tabular}{c}
            \hline
            \multicolumn{1}{|l|}{}                      \\
            \multicolumn{1}{|c|}{\textbf{Goals}}        \\
            \multicolumn{1}{|c|}{\begin{tabular}[c]{@{}c@{}}Integrity\\ Availability\end{tabular}}                \\
            \multicolumn{1}{|l|}{}                      \\
            \multicolumn{1}{|c|}{\textbf{Knowledge}}    \\
            \multicolumn{1}{|c|}{\begin{tabular}[c]{@{}c@{}}White-box\\ Black-box\end{tabular}}                   \\
            \multicolumn{1}{|l|}{}                      \\
            \multicolumn{1}{|c|}{\textbf{Capabilities}} \\
            \multicolumn{1}{|c|}{\begin{tabular}[c]{@{}c@{}}Generic feature-space\\\& Windows Programs\\manipulations\end{tabular}} \\
            \multicolumn{1}{|l|}{}                      \\ \hline
            \multicolumn{1}{l}{\vspace{-10pt}}                       
        \end{tabular}}
        \caption{}
        \label{fig:threat_model}
    \end{subfigure}\hspace{-15pt}
    \begin{subfigure}{0.915\textwidth}
        \centering
        \includegraphics[width=0.9\textwidth]{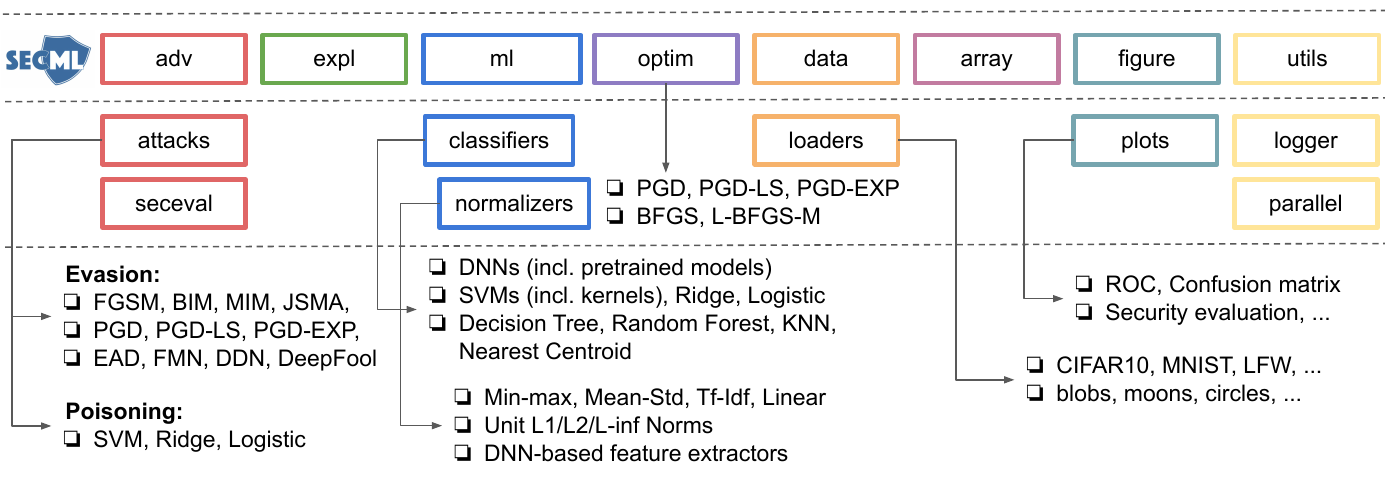}
        \vspace{-3pt}
        \caption{}
        \vspace{-3pt}
        \label{fig:lib-architecture}
    \end{subfigure}
    \label{fig:secml_features_recap}
    \caption{Features available in \secml. On the left, the supported threat models; on the right, the structure of the packages, and the supported strategies.}
\end{figure*}

\section{Software Description}

\secml has a modular architecture oriented to code reuse. 
We have defined abstract interfaces for all components, including loss functions, regularizers, optimizers, classifiers, and attacks. By separating the definition of the optimization problem from the algorithm used to solve it, one can easily define novel attacks or classifiers (in terms of constrained optimization problems) and then use different optimizers to obtain a solution. 
This is a great advantage with respect to other libraries like \cleverhans~\citep{papernot18-technical_report}, \foolbox~\cite{rauber17-arxiv,rauber2017foolboxnative}, and \art~\cite{art18} as, we can switch from white-box to black-box attacks by just changing the optimizer (from a gradient-based to a gradient-free solver), without re-defining the entire optimization problem.

\secml integrates different components via well-designed wrapper classes.
We integrate several attack implementations from \cleverhans and \foolbox, by extending them to also track the values of the loss function and the intermediate points optimized during the attack iterations, as well as the number of function and gradient evaluations. 
This is useful to debug and compare different attacks, e.g., by checking their convergence to a local optimum and properly tuning their hyperparameters (e.g., step size and number of iterations).
\secml supports DNNs via a dedicated \pytorch wrapper, which can be extended to include other popular deep-learning frameworks, like \tensorflow and \keras, and it natively supports \sklearn classifiers as well.
This allows us to run attacks that are implemented in \cleverhans and \foolbox to universally run on both \pytorch and \sklearn models.

\myparagraph{Main packages.} The library supports different threat models (Fig.~\ref{fig:threat_model}), and it is organized in different packages (Fig.~\ref{fig:lib-architecture}). 
The \adv package implements different adversarial attacks and provides the functionalities to perform security evaluations. It includes target and untargeted evasion attacks provided by \cleverhans and \foolbox as well as our implementations of evasion and poisoning attacks~\cite{biggio18}.
These attacks target the feature space of a model, regardless of its domain of application (e.g. attacking Android malware detection\footnote{\url{https://github.com/pralab/secml/blob/master/tutorials/13-Android-Malware-Detection.ipynb}}).
Hence, developers can encode ad-hoc manipulations that correspond to perturbations in the input space of a targeted domain (e.g. Windows programs~\citep{demetrio2021secml}), and then use the algorithms included inside \secml to optimize the attack. 
All the included attacks are listed in Fig.~\ref{fig:lib-architecture}.

The \ml package imports classifiers from \sklearn and DNNs from \pytorch. 
We have extended \sklearn classifiers with the gradients required to run evasion and poisoning attacks, which have been implemented analytically.
Our library also supports chaining different modules (e.g., scalers and classifiers) and can automatically compute the corresponding end-to-end gradient via the chain rule.
The \explanation package implements the feature- and prototype-based explanation methods~\cite{ribeiro16,sundararajan17-icml,koh2017understanding}.
The \optim package provides an implementation of the projected gradient descent (PGD) algorithm, and a more efficient version of it that runs a bisect line search along the gradient direction (PGD-LS) to reduce the number of  gradient evaluations~\cite{demontis19-usenix}. 
Finally, \data provides data loaders for popular datasets, integrating those provided by \sklearn and \pytorch; \texttt{array} provides a higher-level interface for both dense (\numpy) and sparse (\scipy) arrays, enabling the efficient execution of attacks on sparse data representations; \pkfigure implements some advanced plotting functions based on \matplotlib (e.g., to visualize and debug attacks); and \textit{utils} provides functionalities for logging and parallel code execution.
We also provide a model zoo, available on GitHub as well,\footnote{\url{https://github.com/pralab/secml-zoo}} that contains pre-trained models to rapidly test newly implemented attacks and utilities.

\begin{table*}[]
\vspace{2cm}
\centering
\begin{tabular}{cccccccccccc}
\multicolumn{1}{l}{} &
  \multicolumn{1}{l}{\begin{rotate}{60}\scriptsize DL frameworks support\end{rotate}} &
  \multicolumn{1}{l}{\begin{rotate}{60}\scriptsize\sklearn support\end{rotate}} &
  \multicolumn{1}{l}{\begin{rotate}{60}\scriptsize Built-in attack algorithms\end{rotate}} &
  \multicolumn{1}{l}{\begin{rotate}{60}\scriptsize Wraps adversarial frameworks\end{rotate}} &
  \multicolumn{1}{l}{\begin{rotate}{60}\scriptsize Dense / Sparse data support\end{rotate}} &
  \multicolumn{1}{l}{\begin{rotate}{60}\scriptsize Security evaluation plots\end{rotate}} &
  \multicolumn{1}{l}{\begin{rotate}{60}\scriptsize Attack loss inspection plots\end{rotate}} &
  \multicolumn{1}{l}{\begin{rotate}{60}\scriptsize Loss separated from Optimizer\end{rotate}} &
  \multicolumn{1}{l}{\begin{rotate}{60}\scriptsize Explainability\end{rotate}} &
  \multicolumn{1}{l}{\begin{rotate}{60}\scriptsize Model zoo\end{rotate}} &
  \multicolumn{1}{l}{\begin{rotate}{60}\scriptsize Comprehensive tutorials\end{rotate}} \\
  \normalsize
\textbf{\secml}      & \checkmark & \checkmark & \checkmark & \checkmark &  \checkmark & \checkmark & \checkmark & \checkmark & \checkmark & \checkmark & \checkmark \\ \hline
\foolbox    & \checkmark & \xmark     & \checkmark & \xmark     & \xmark     & $\sim$     & \xmark & \xmark  & \xmark     & \checkmark & $\sim$     \\ \hline
\art        & \checkmark & \checkmark & \checkmark & \xmark     & \xmark     & \xmark     & \xmark     & \xmark   & \xmark    & \xmark & \checkmark \\ \hline
\cleverhans & \checkmark & \xmark     & \checkmark & \xmark     & \xmark     & \xmark     & \xmark     & \xmark   & \xmark    & \xmark & \xmark     \\ \hline
\end{tabular}
\caption{Comparison of \secml and the other adversarial machine learning libraries. We show whether the library offer full (\checkmark), partial ($\sim$) or no (\xmark) support of a particular feature.}
\label{tab:tools-comparison}
\end{table*}

We recap the main functionalities of \secml in Table~\ref{tab:tools-comparison}, where we also compare it with other relevant libraries. Notably, our library is the sole that provides attack loss inspection plots to choose the appropriate attacks' hyperparameters, and security evaluation plots, to ease the complexity of assessing the robustness of machine learning models.
Moreover, the features offered by \secml are not only related to attacking machine learning models, but they are gathered to elevate \secml as a complete tool for attacking, inspecting, and assessing the performances of machine learning models.

\myparagraph{Testing and documentation.} We have run extensive tests on macOS X, Ubuntu 16.04, Debian 9 and 10, through the GitHub Actions infrastracture,\footnote{\url{https://docs.github.com/en/actions/automating-builds-and-tests}} with also experimental support on Windows 10 platforms.
The user documentation is available at \url{https://secml.readthedocs.io/en/v0.15/}, along with a basic developer guide detailing how to extend the \ml package with other classifiers and deep-learning frameworks. 
The complete set of unit tests is available in our repository. 
%Many Python notebooks are also available, and they collect a comprehensive view of the functionalities available inside \secml, with step-by-step tutorials on how to use them.
A comprehensive view of the functionalities available in \secml is included in tutorials available as Jupyter notebooks.

\section{Impact}

We now offer two examples extracted from \secml to showcase its impact: evasion attacks against DNNs, and a poisoning attack against an SVM.

\myparagraph{Evasion attacks.}
We show here how to use \texttt{secml} to run different evasion attacks against ResNet-18, a DNN pretrained on ImageNet, available from \torchvision.
This example demonstrates how \texttt{secml} enables running \cleverhans attacks (implemented in \tensorflow) against \pytorch models. 
Our goal is perturbing the image of a race car to be misclassified as a tiger, using the $\ell_2$-norm targeted Carlini-Wagner (CW) attack (from \cleverhans), the $\ell_2$ PGD attack implemented in \secml, and PGD-patch, where the attacker can only change the pixels corresponding to the license plate~\cite{melis17-vipar}.

\begin{figure}[t]
\centering
    \includegraphics[width=.19\textwidth]{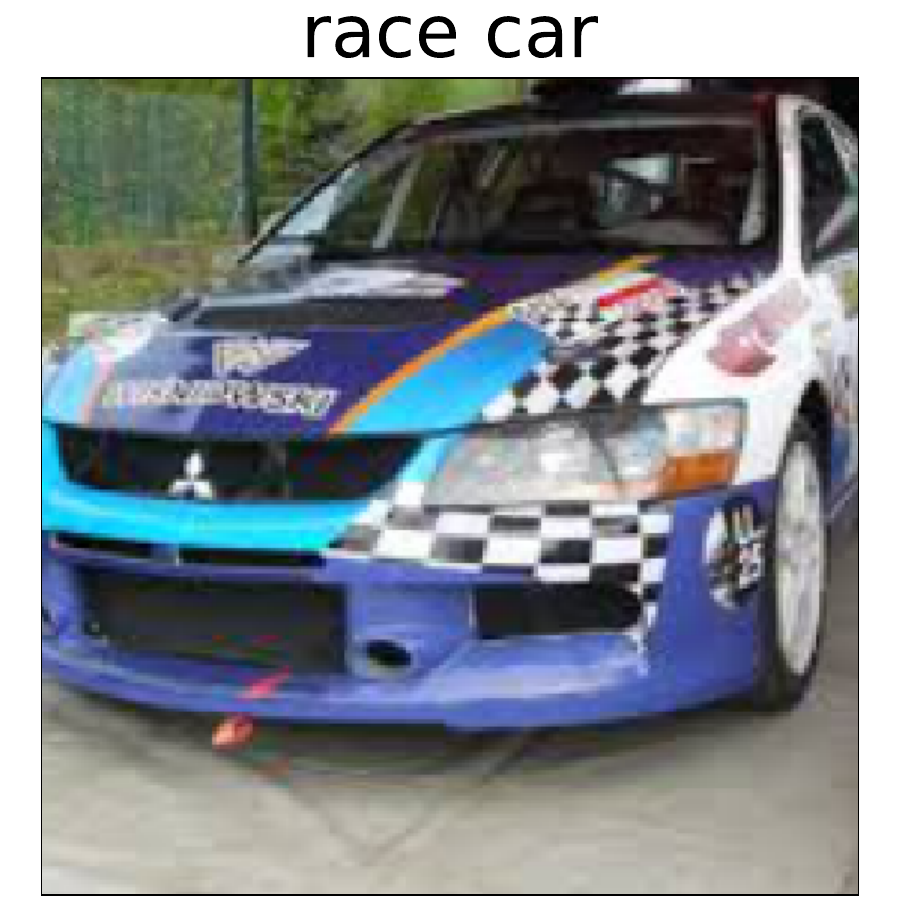} 
    \includegraphics[width=.19\textwidth]{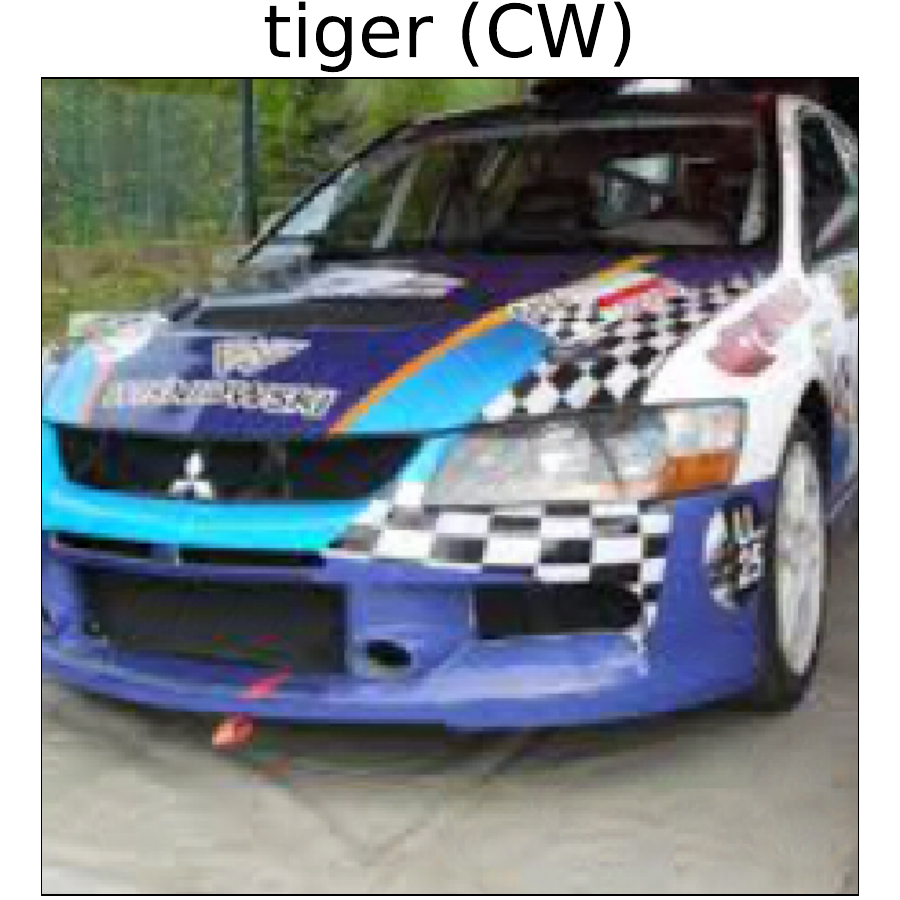} 
    \includegraphics[width=.19\textwidth]{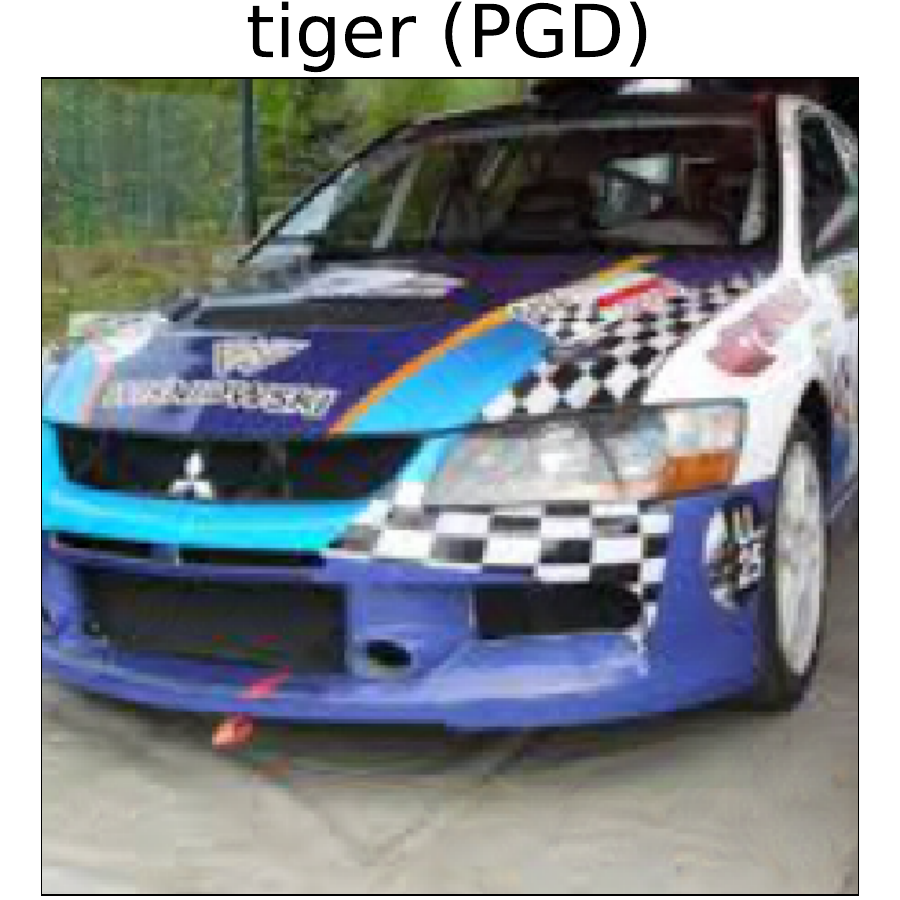}
    \includegraphics[width=.19\textwidth]{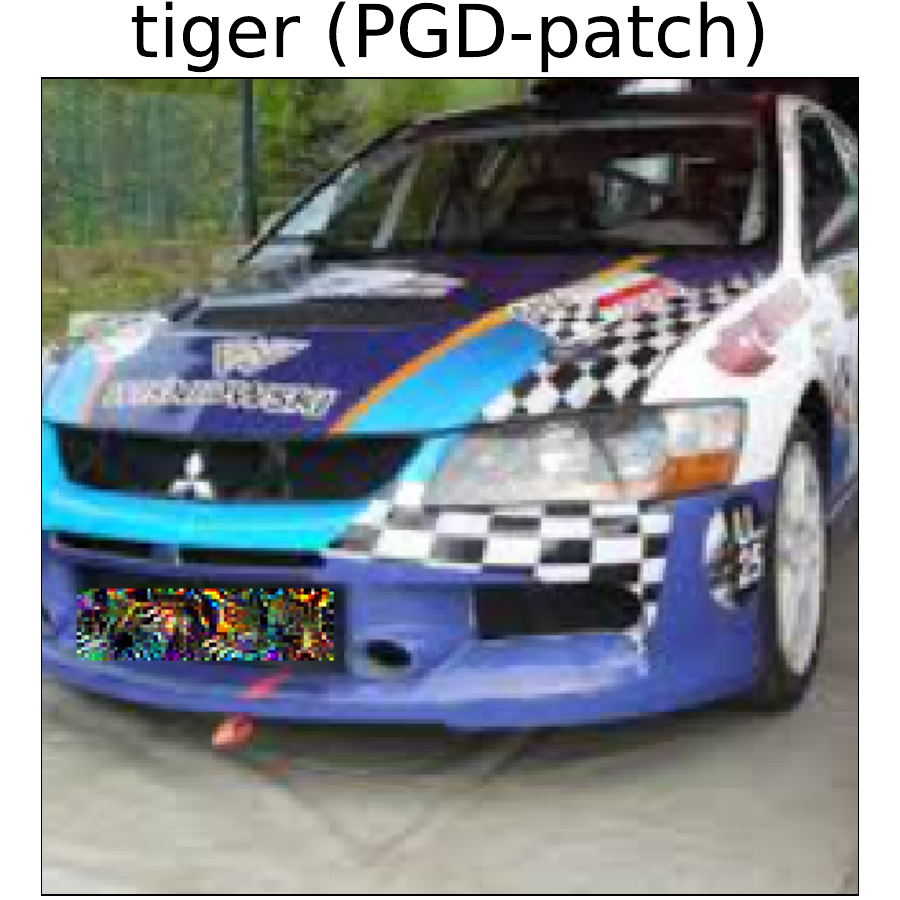}
    \includegraphics[width=.19\textwidth]{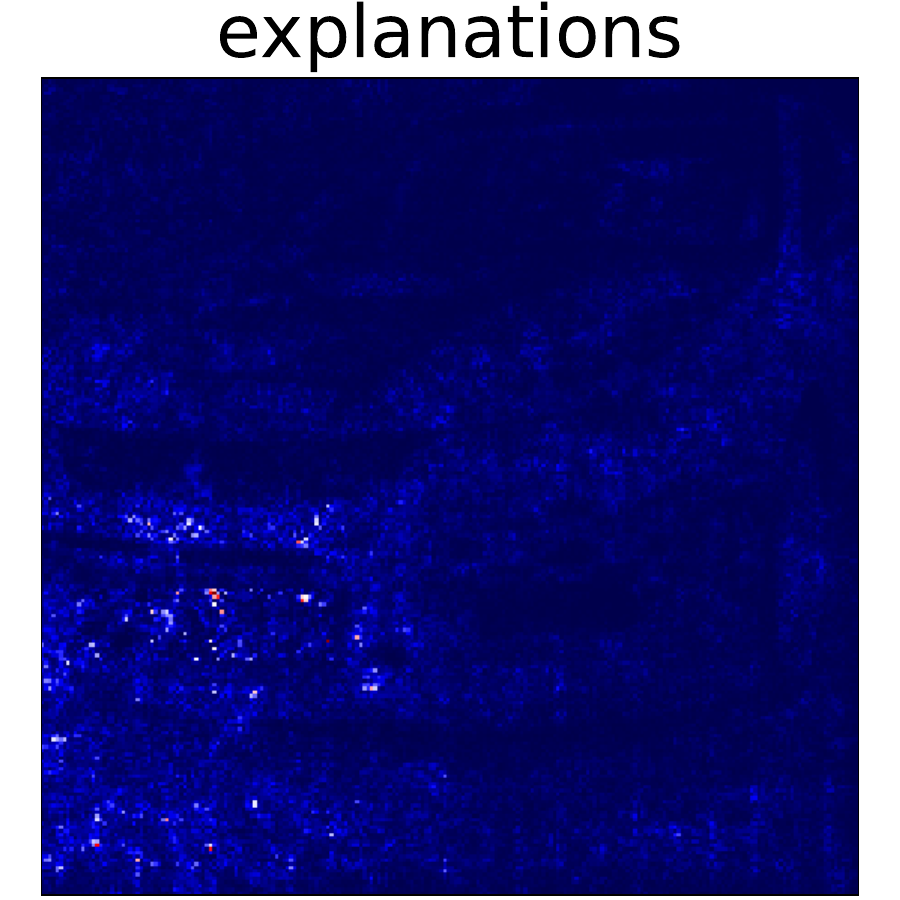}
    \vspace{-6pt}
    \caption{Adversarial images (CW, PGD, and PGD-patch) representing a \emph{race car} misclassified as a \emph{tiger}. For PGD-patch, we also report explanations via integrated gradients.}
    \label{fig:attacks}
\end{figure}

\begin{figure}[t]
\centering %.4
    \includegraphics[width=.47\textwidth]{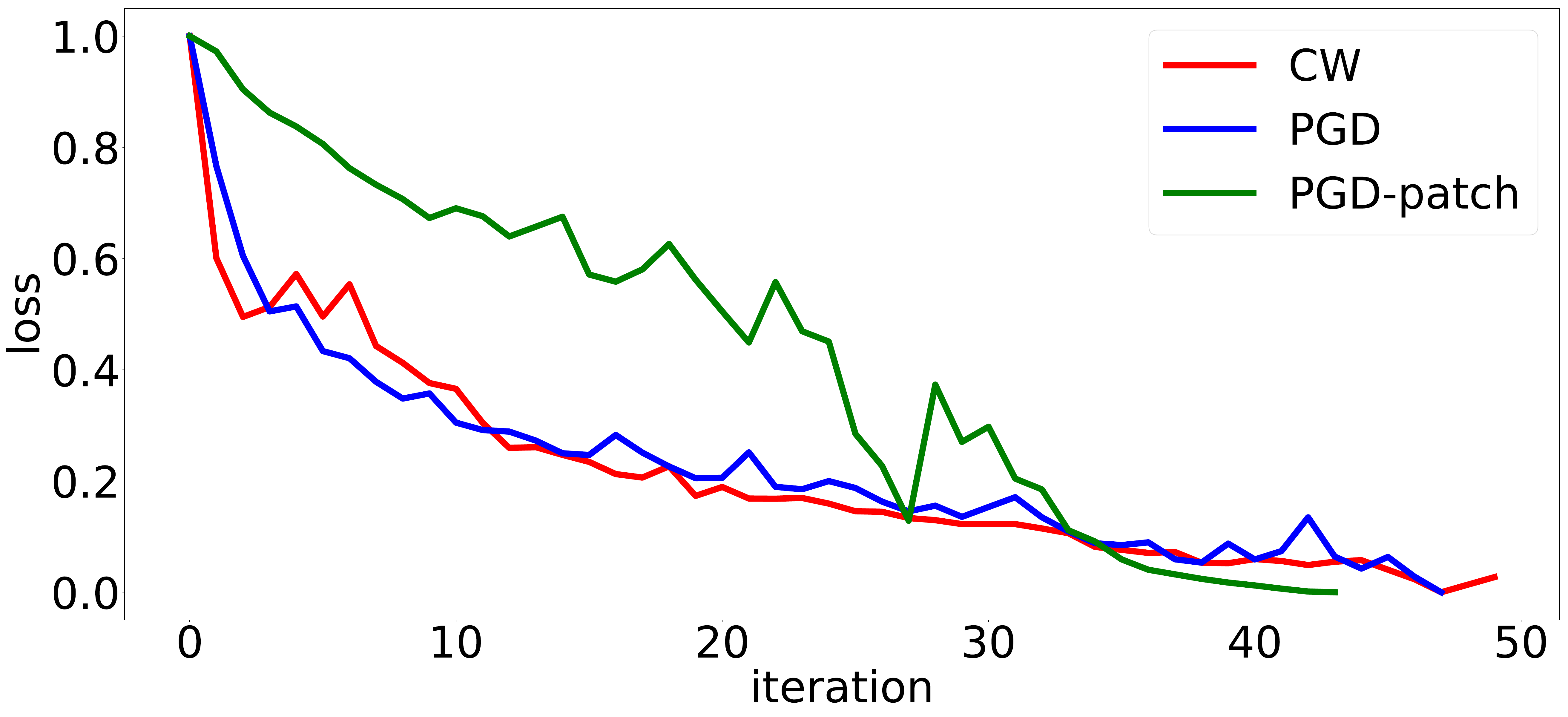} 
    \includegraphics[width=.47\textwidth]{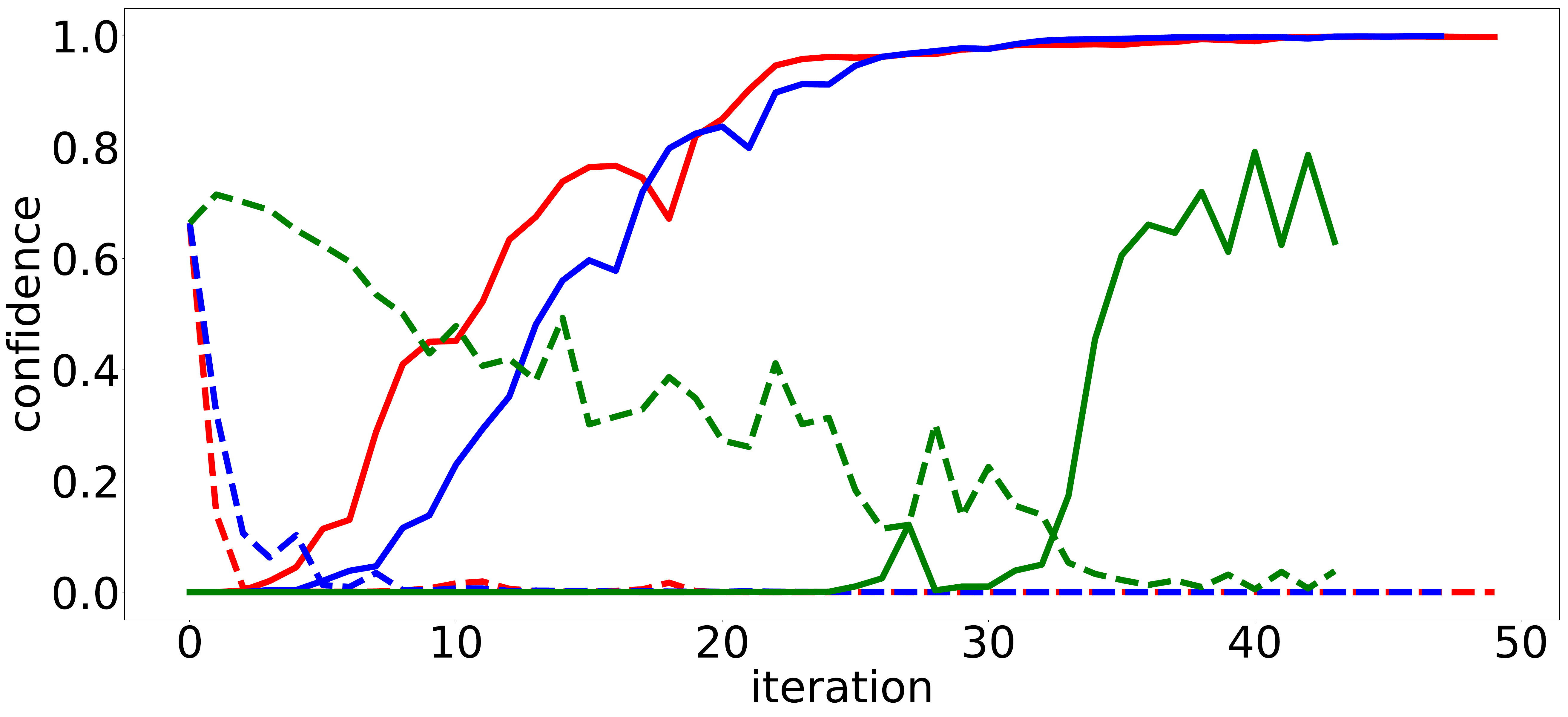}
    \vspace{-3pt}  
    \caption{Attack optimization. \emph{Left}: loss minimization; \emph{Right}: confidence of source class (\emph{race car}, dashed lines) vs confidence of target class (\emph{tiger}, solid lines), across iterations.}
    \label{fig:debugging-plots}
\end{figure}

\begin{figure*}[ht]
    \centering
    \includegraphics[width=0.95\textwidth]{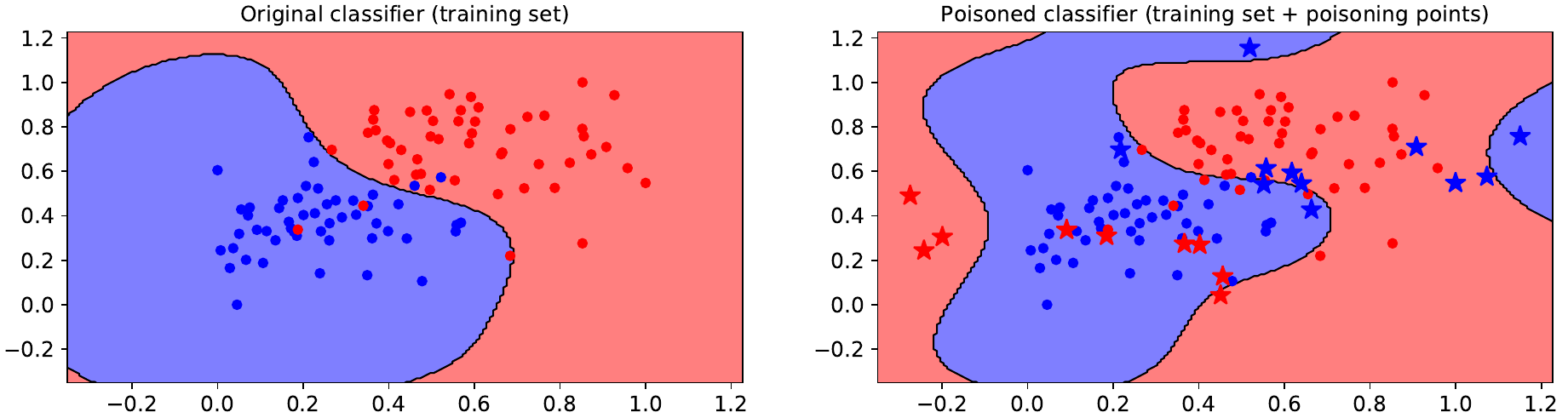}
    \caption{Poisoning attacks against an SVM implemented with \sklearn. The poisoning data (denoted with $\star$ in the right plot) induce the model to learn a worse decision boundary.}
    \label{fig:poisoning_example}
\end{figure*}

We run all the attacks for $50$ iterations, we set the confidence parameter of the CW attack $\kappa=10^6$ to generate high-confidence misclassifications, and $c=0.4$, yielding an $\ell_2$ perturbation size $\epsilon=1.87$. 
We bound PGD to create an adversarial image with the same perturbation size. 
For PGD-patch, we do not bound the perturbation size for the pixels that can be modified.

The resulting adversarial images are shown in Fig.~\ref{fig:attacks}. 
For PGD-patch, we also highlight the most relevant pixels used by the DNN to classify this image as a tiger, using the \emph{integrated gradients} explanation method.
The most relevant pixels are found around the perturbed region containing the license plate, unveiling the presence of potential adversarial manipulations.

We also visualize the performances of the attack in Fig.~\ref{fig:debugging-plots}.
The leftmost plot shows how the attack losses (scaled linearly in $[0,1]$ to enable comparison) iteration-wise, while the rightmost plot shows how the confidence assigned to class \emph{race car} (dashed line) decreases in favor of the confidence assigned to class \emph{tiger} (solid line) for each attack, across different iterations.
By inspecting them, we can understand if these attacks have been correctly configured.
For instance, by looking at the loss curves on the left, we can understand if the attacks reached convergence or not, thus facilitating tuning of either the step size or the number of iterations.
Also, by looking at the plot on the right, it is clear that all the attacks are successful since the confidence of the target class exceeds the score of the original one.

%Finally, in Fig.~\ref{fig:debugging-plots}, we report some plots to better understand the attack optimization process. \mpcomment{qui va messo come si usano le curve o come si interpretano} The leftmost plot shows how the attack losses (scaled linearly in $[0,1]$ to enable comparison) are minimized while the attacks iterate. The rightmost plot shows how the confidence assigned to class \emph{race car} (dashed line) decreases in favor of the confidence assigned to class \emph{tiger} (solid line) for each attack, across different iterations. We have found these plots particularly useful to tune the attack hyperparameters (e.g., step size and number of iterations), and to check their converge to a good local optimum.
%We firmly believe that such visualizations will help avoid common pitfalls in the security evaluation of learning algorithms, facilitating understanding and configuration of the attack algorithms.
\myparagraph{Poisoning attacks.}
We also show the effect of a poisoning attack provided by \secml applied to an SVM classifier implemented in \sklearn. The experimental setting and code are available in one of our tutorials on GitHub\footnote{\url{https://github.com/pralab/secml/blob/master/tutorials/05-Poisoning.ipynb}}.
Results of the successful attack are represented in Fig.~\ref{fig:poisoning_example}., highlighting the flexibility of \secml in applying different strategies to third-party models as well, without the need of customizing them on a particular framework.

\section{Conclusions and Future Work}
The \secml project was born more than five years ago and we open-sourced it in August 2019. Thanks to an emerging community of users and developers from our GitHub repository, we firmly believe that \secml will soon become a reference tool to evaluate the security of machine-learning algorithms. We are constantly working to enrich it with new functionalities, by adding novel defenses, wrappers for other third-party libraries, and more pretrained models to the \secml zoo.

\section*{Acknowledgements}
This work has been partly supported by the PRIN 2017 project RexLearn, funded by the Italian Ministry of Education, University and Research (grant no. 2017TWNMH2); by the project ALOHA, under the EU's H2020 programme (grant no. 780788); and by the project TESTABLE (grant no. 101019206), under the EU’s H2020 research and innovation programme.

\end{document}